# Parallel and Streaming Wavelet Neural Networks for Classification and Regression under Apache Spark

*Eduru Harindra Venkatesh[1,2], Yelleti Vivek[1], Vadlamani Ravi[1*] and*

*Orsu Shiva Shankar[1]*

[1]Center for Artificial Intelligence and Machine Learning,
Institute for Development and Research in Banking Technology,
Castle Hills Road #1, Masab Tank, Hyderabad-500076, India
[2]School of Computer science and Information Sciences, University Of Hyderabad,
Gachibowli, Hyderabad-500046, India
harindravenkatesh@gmail.com; yvivek@idrbt.ac.in; vravi@idrbt.ac.in; shivaorsu96@gmail.com

**Abstract**

Wavelet neural networks (WNN) have been applied in many fields to solve regression as well as classification problems. After the advent of big data, as data gets generated at a brisk pace, it is imperative to analyze it as soon as it is generated owing to the fact that the nature of the data may change dramatically in short time intervals. This is necessitated by the fact that big data is all pervasive and throws computational challenges for data scientists. Therefore, in this paper, we built an efficient Scalable, Parallelized Wavelet Neural Network (SPWNN) which employs the parallel stochastic gradient algorithm (SGD) algorithm. SPWNN is designed and developed under both static and streaming environments in the horizontal parallelization framework. SPWNN is implemented by using Morlet and Gaussian functions as activation functions. This study is conducted on big datasets like gas sensor data which has more than 4 million samples and medical research data which has more than 10,000 features, which are high dimensional in nature. The experimental analysis indicates that in the static environment, SPWNN with Morlet activation function outperformed SPWNN with Gaussian on the classification datasets. However, in the case of regression, the opposite was observed. In contrast, in the streaming environment i.e., Gaussian outperformed Morlet on the classification and Morlet outperformed Gaussian on the regression datasets. Overall, the proposed SPWNN architecture achieved a speedup of 1.32-1.40.

## 1. Introduction

In recent times, wavelet analysis [6] had become popularized in handling various tasks such as analyzing time series data [2] and images [3], speech recognition [5,9], computer vision [24], financial forecasting [22] etc., Wavelet analysis is a mathematical tool used in various areas of research. Wavelets describe the time series and are represented by local information such as frequency, duration, intensity, and time position, and by global information such as the mean states over different time periods. Both global and local information is needed for the correct analysis of a signal. The wavelet transform (WT) is a generalization of the Fourier transform (FT) and the windowed Fourier transform (WFT). The neural network which is inspired by the wavelet analysis is regarded as a Wavelet Neural Network (WNN), [14, 28] which is a combination of feed forward neural network and wavelet analysis. WNN provides the following advantage over other neural networks (NNs) it takes less training time when compared to a multilayer perceptron network [23] without sacrificing the performance.

    In today's world, almost every sector has a very huge amount of data which is formally regarded as Big data. Analysing such large data sets is more complex than standard relational databases in terms

---

[1*] Corresponding Author

of capturing, handling and processing data quickly. The main characteristics of Big data are viz., Volume, Velocity, Variety, Veracity, and Value (5V's) [11]. Such Big Data may not be processed easily in a single system. This gave origin to a new paradigm of computing i.e., Distributed computing. This is the technique of executing a complex computation on a number of distinct processors that may exist on the same system, on many computers in the same network, or different machines [10].

To capture and analyse Big Data, several tools were proposed. Among them, the most popular tools are Apache Hadoop and Apache Spark. Apache Hadoop is a framework that enables distributed processing of large data sets across computer groups using MapReduce [4] as the programming model. It is designed to scale from a single server to thousands of machines, each machine providing local computing and storage. The framework itself does not rely on hardware to provide high availability but is designed to detect and handle application layer failures, thereby providing high availability services on a group of computers, each of which may be prone to failure. On the other hand, Apache Spark is a unified analytics engine for large-scale data processing. The main abstraction provided by Spark is the Resilient Distributed Data Set (RDD), which is a collection of partitioned items that can be operated in parallel across cluster nodes. Further, it provides high-level APIs in Java, Scala, R, and Python, and an optimization engine that supports general execution graphs. In addition, Spark also supports a broad set of advanced tools, which includes Spark SQL for SQL, structured data processing, MLlib for machine learning, GraphX for graph processing, and structured flows for incremental computation. These parallel and distributed solutions can decrease the time [8].

As we discussed earlier, Velocity is one of the important characteristic of Big Data, where the data is continuously generated by various data sources. Such data is called Streaming data. This data is being generated from thousands of data sources every second and must be processed and analyzed as quickly as possible. These streaming demands a new paradigm of solutions to handle the data which comes from time-to-time [19]. For example, Google search results, posts shared in various social media platforms, stock market trends etc.,. We can get deeper insights or extract useful information at an early phase itself, rather than waiting for the whole data. Hence, there is an utmost need to design the streaming applications to analyzing the large volumes of streaming data on-time. Creating models for the desired task is challenging and certain modifications need to be done in the extant methods to achieve the faster analysis on-time without compromising on the performance of the model [20].

Interestingly, Spark also supports streaming which is of different types. Spark Streaming is an extension of Spark's core API, which enables scalable, high-performance, and fault-tolerant stream processing of real-time data streams. Structured Streaming is a fault-tolerant and scalable stream processing engine built on the Spark SQL engine. We can express our stream computing in the same way as batch computing on static environment. The Spark SQL engine will be responsible for running it continuously in increments and updating the final result as streaming data continues to arrive. Spark provides a high-level abstraction called discretized stream (DStreams), which represents a continuous data stream. The data can be ingested from many sources (such as Kafka, Kinesis, or TCP sockets) and can be processed using complex algorithms represented by high-level functions (such as map, reduce, join, and window). Further details of Dstreams and its operations will be discussed in-detail in latter sections.

Optimization of neural networks is a crucial task to reduce losses and for providing accurate results for the desired task. Gradient descent is widely used and has been the most common approach for optimizing neural networks [17, 18]. But, when applied to large amount of data, this often consumes too much time which is not desirable. Hence, a method called Stochastic gradient descent (SGD) is proposed [32-33], is one such optimization which updates the parameters for each training sample. SGD performs one update at a time and is much faster. Basically, SGD is proven to be handling this reducing the computational complexity without compromising on the performance of the model. This is well proven especially at high dimensional datasets. There are several variants of SGD were proposed in the literature [32-33]. Among them, SGD with mini batch gradient is proven to be effective hence this made us to adopt it in the proposed parallel model.

The above discussion motivated us to design a robust and scalable WNN algorithm to make it suitable for large datasets and high dimensional problems. Hence, we proposed scalable, parallel WNN and named it SPWNN under Apache Spark environment. The proposed SPWNN is suitable to solve both the classification and regression problems. Further, as we are dealing with Big Data, we adopted the SGD optimization in the proposed parallel model. In addition, we also proposed a online version of SPWNN to handle the streaming data.

The major contributions in the current study are as follows:

- We developed and designed a scalable parallel WNN and named it SPWNN under the Spark framework.
- SPWNN is developed under both static and streaming environments.
- The performance of the SPWNN is analysed to solve both the classification and regression problems.
- For faster optimization, SPWNN is fused with the SGD optimization principles to make it a light-weight model.

The rest of the current research study is organized as follows: Section 2 contains the literature review. Section 3 contains the background theory extensively used for this research. Section 4 contains the proposed methodology and algorithms. Section 5 contains the experiments conducted and the results yielded. Section 6 contains the conclusions to this research.

## 2. Background

In this section, we will discuss about the background theory relevant to the current study.

### 2.1 Wavelet Neural Network

Wavelets are very popular and have been applied in many fields of research. The expansion of the wavelet series shows the location of the time frequency of a given signal. By combining wavelets with neural networks (NN), wavelet neural networks (WNN) have been developed [27]. Although neural networks have important features like learning, generalization, and parallel computing, they require a large number of neurons in the hidden layer of the network to solve the function learning problem and cannot converge quickly. In addition to the good characteristics of NN, WNN can also converge quickly and provide high precision while reducing the size of the network, because the time-frequency positioning characteristics of the waves allow the shortest time to converge to their global maximum.

The WNN model combines the advantages of discrete wavelet transform and neural network processing to achieve strong nonlinear approximation capabilities and has been successfully applied to function prediction, modelling, and approximation. The Wavelet neural network (WNN) architecture is based on the multilayer perceptron (MLP). In the case of WNN, the discrete wavelet function is issued as the node activation function.

Wavelet networks are usually in the form of a three-layer network. The bottom layer represents the input layer, the middle layer is the hidden layer, and the top layer is the output layer. In the input layer, the explanatory variables are entered into the wavelet network. The hidden layer is made up of hidden units (HU). The hidden units, also called wave elements, are similar to neurons in the classical sigmoid neural network. In the hidden layer, the input variable becomes an expanded and translated version of the parent wavelet. Finally, in the output layer, the approximate value of the target value is estimated.

## 2.2 Parallel Stochastic Gradient Descent

In general, SGD picks data points at random from a set of data. However, in the big data environment, the data is divided into multiple partitions. The parallel SGD [31] works in the following way: (i) pick random data points at every salve node & compute the weights. (ii) collects all the weights in the master node. (iii) Aggregates the weights and the average of thus collected weights from various slaves is computed. (iv) Thus aggregated weights were broadcasted to all of the workers to further training. This process is repeated for user defined number of epochs. It's worth noting that this model aggregation and redistribution is how this implementation used several computers to train a single neural network model in a coordinated manner, as well as how this implementation is synchronous.

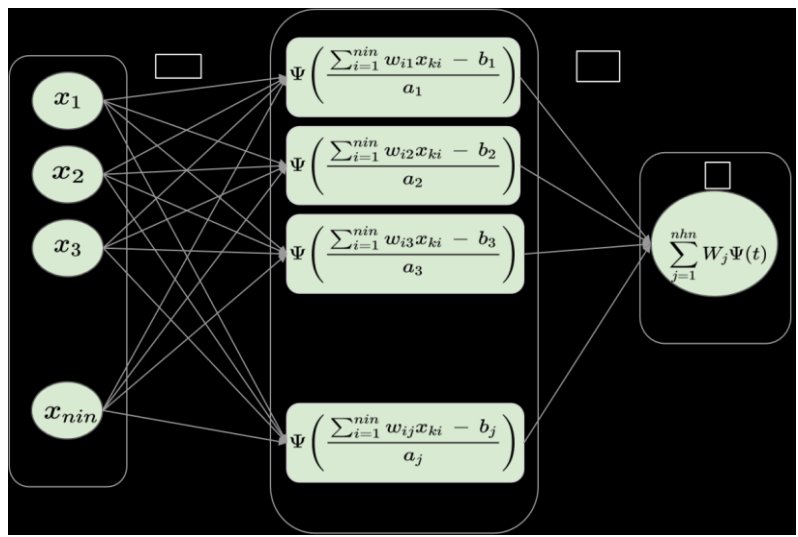

Fig. 1: Wavelet Neural Network

## 2.3 RDD

RDD supports two operations: (i) Transformations, which create a new data set from an existing data set, and (ii) Actions, which return a value to the controller/driver after running a calculation or performing a computation on the dataset. RDD works differently in local and cluster. In order to run jobs, Spark divides the processing of RDD operations into tasks, and each task is run by an executor. Before execution, the Spark computing task's closure. The closure is those variables and methods that must be visible for the executor to perform calculations in the RDD.

## 2.4 Discretized Stream

Spark Streaming's underlying abstraction is the Discretized Stream, or DStream depicted in Fig. 2. It depicts a continuous stream of data, either the source's input data stream or the processed data stream produced by transforming the input stream. Spark's abstraction of an immutable, distributed dataset is a DStream, which is represented internally by a continuous succession of RDDs. A DStream's RDDs each hold data from a certain interval. Any operation applied on a DStream translates to operations on the underlying RDDs. These underlying RDD transformations are computed by the Spark engine. The DStream operations hide most of these details and provide the developer with a higher-level API for convenience.

File Streams are used to read data from a file or a local location and can be used as streaming data. Here we do not require any receiver to collect the data, so there is no need to allocate any cores for receiving data. But unfortunately, this API is not available for Python API. So, we use Queue Streams in order to replicate this streaming from a file source.

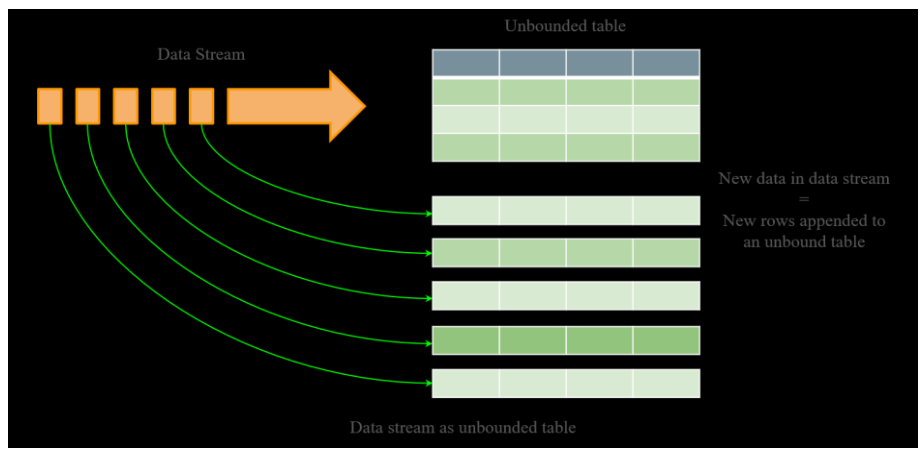

Fig. 2: DStreams

**2.5 Queue Streams & Window Operations**

Spark provides an API to construct DStreams from a series of RDDs called Queue Stream. With this one can create an input stream from a queue of RDDs or a list, this queue Stream can process either one at a time or all at once.

Spark Streaming also provides windowed operations with which we perform various computations on the data. It also allows us to apply transformations over a sliding window of data. The source RDDs that fall within the window are combined and operated upon to produce the RDDs of the windowed DStream as depicted in Fig. 3. Any window operations need two parameters which are:

- Window Length - The duration of the window.
- Sliding Interval - The interval at which the window operation is performed.

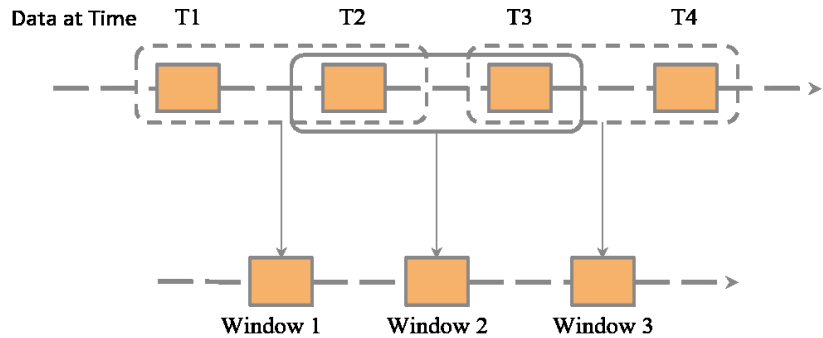

Fig. 3: Sliding Window Operation on Dstreams

## 3. Literature Survey

This section starts by discussing the works related to WNN, followed by SGD and then concludes with the discussion on streaming.

**3.1 Literature Survey related to WNN**

Wavelet neural networks have been proposed in the year 1992 [28]. These WNNs are used for various machine learning problems such as classification, and regression etc., One such problem statement is cost estimation using WNN [13]. The authors of [13] implemented two variants of WNN by using two different wavelet functions viz., Morlet and Gaussian. A detailed comparative analysis is conducted with the traditional SVM model and MLP and others where WNN outperformed the state-of-art models. In addition, WNN was also implemented on time series prediction [2] where the authors incorporated a linear layer used between hidden and output layers resulting in a variant of WNN. This approach allows the advantage of requiring fewer wavelets than the traditional WNN. The usefulness of the proposed approach was demonstrated by the results for the time-series prediction problem. Rainfall prediction was also done using the WNN [15]. The results state that WNN model outperformed ANN models in monthly rainfall prediction tasks. WNN model achieved over 94 percent of the efficiency index, whereas ANN models achieved 64 percent only. This proves the importance of WNN and its applicability in various case-studies. Also, WNN is also quite popular in handling dynamic systems [25]. Authors from [25], proved that fuzzy based WNN has better generalizability and able to produce IF then rules, where Gaussian activation function is used in the IF part.

**3.2 Literature Survey related to SGD**

Stochastic gradient descent (SGD) is the most commonly utilized optimizing technique because it's simple to use and easy to adapt for large volumes of data [26]. The Stochastic Approximation Method was first proposed in the year 1951 [16]. As a result, parallelizing SGD and developing SGD variants has sparked a lot of interest.

One interesting variation of SGD is mini batch stochastic gradient descent. The typical strategy of a mini-batch is to sample uniformly at every stage, leading often too large a variance. Research has been done using stratified sampling for mini-batch SGD [30]. As part of the research results were compared with dynamic and stratified sampling and also with uniform sampling. It was proven that the convergence rate has been substantially improved.

Parallelizing SGD is necessary for data which takes hours to read content from the disk as computation becomes infeasible. Simu Parallel SGD [31] is one such algorithm which proposes a data parallel algorithm for optimization of a network. The methodology claims to involve very less utility of I/O. The algorithm can be implemented using Map Reduce and thus gains the property of handling fault tolerance. The authors of [31] claimed to produce solutions with a reduced time when parallelizing is applied. SGD has also been implemented on HPCC systems which use the synchronous data parallel SGD technique [12]. The results show a consistent decrease in the time taken for model training in terms of dataset and cluster size. It was also observed that training time when a new node is added to a cluster is relative to both computational power and communication overhead. The approach works well with various dataset sizes and can scale over several cluster sizes.

### 3.3 Literature Survey related to Streaming models

Computation of Streaming data involves handling large amounts of data generated at a rapid pace from various sources with low latency. The assumption of streaming data is that the potential worth of data depends on the freshness of the data. Streaming allows businesses to analyze and respond in real time. But handling this type of streaming data is difficult and it is necessary to develop parallel and scalable architectures. Real time streaming big data (RT-SBD) [19] processing engine was proposed which employs the methodology of using multicore processors for the processing of real time streaming data. The model uses a multiprocessing environment where each query is executed in parallel. A hybrid clustering scheduling algorithm is used for query assignment for the processors in the cluster.

The Research was also done using the K-means clustering algorithm for streaming data using Spark and Hadoop [21]. The work involves collecting data from slack, an application which produces real time data and sparks streaming was used for data collection. The data is clustered based on languages and this data is translated into English. Using spark streaming has facilitated the work with fast processing and promising results. It is also stated that adding more Spark servers in a distributed fashion will lead to increased scalability.

Spark's remarkable performance was confirmed by previous works [1]. The authors of [1] also present results of Spark with different streaming inputs like Kafka, TCP and file streaming. Harmonic IO is also implemented for comparison analysis. The work also demonstrated the necessity of selecting a stream source integration that is adequate for the message size and performance requirements.

## 4. Proposed Methodology

This section begins with a discussion on proposed parallel WNN algorithms for regression and classification with the distinction involved. Later, how the WNN is extended for streaming problems is also discussed.

### 4.1 SPWNN for Classification and Regression

SPWNN is used to perform regression analysis to predict the target variable ($V_k$) based on input variables in a distributed environment using stochastic gradient descent (SGD). The algorithm of SPWNN is given in Algorithm 1. The major reason behind using the SGD is to decrease the time-complexity.

SPWNN for any machine learning problem i.e., for both the regression/classification starts by initializing the required parameters such as Dilation, Translation parameters, and Weight vectors which are randomly initialized ( both the weights between the input layer and hidden layer and also the weights between the hidden layer and output layer), number of hidden nodes (*nhn*), epochs (*N*), learning rate (*lr*), momentum (*m*) and batch size (*bs*). After initializing the parameters, all the parameters are broadcasted to all the nodes.

Once, the parameters are initialized, the mapper function is called from the driver. The data is divided into multiple partitions and on each partition, the mapper function is executed. Here, we are using the SGD optimization, hence in each partition, the mapper function is applied only on a single random point. By calling mapper, forward propagation with the SGD optimization is performed. It is to be noted that all of the partitions will get an identical model. As we had adopted mini batch SGD optimization, which is a popular and efficient method of SGD. By using Eq. (1) the $V_k$ is computed.

$$V_k = \sum_{j=1}^{nhn} W_j f\left(\frac{\sum_{i=1}^{nin} w_{ij} x_{ki} - b_j}{a_j}\right) \tag{1}$$

The $V_k$ is dependant on the underlying activation function. In our approach, two different activation functions were studied (i) Morlet function (refer to Eq. (2)); (ii) Gaussian function (refer to Eq. (3)). Thereafter, the weight were updated according to $V_k$.

$$f(t) = \cos(1.75t) \exp\left(-\frac{t^2}{2}\right) \tag{2}$$

$$f(t) = \exp(-t^2) \tag{3}$$

The above process is performed parallel on all of the data partitions. All the gradients and computed weights in each data partition is performed parallely among worked nodes are collected at the master node. This is achieved by utilizing the steps as given in Eq. (4) – (7). The learning rate $\eta$ and $\alpha$ is the momentum are passed as parameters. These series of steps comprising of updating weights is performed in parallel in all the data partitions.

$$\Delta W_j(t+1) = -\eta \frac{\partial E}{\partial W_j(t)} + \alpha \Delta W_j(t) \tag{4}$$

$$\Delta w_{ij}(t+1) = -\eta \frac{\partial E}{\partial w_{ij}(t)} + \alpha \Delta w_{ij}(t) \tag{5}$$

$$\Delta a_j(t+1) = -\eta \frac{\partial E}{\partial a_j(t)} + \alpha \Delta a_j(t) \tag{6}$$

$$\Delta b_j(t+1) = -\eta \frac{\partial E}{\partial b_j(t)} + \alpha \Delta b_j(t) \tag{7}$$

After the completion of the previous steps, thus trained parameters i.e., *w,W,a,b* were collected back to the driver. Then the average of these is computed and then the error is also computed. In the case of regression, we have used mean squared error (MSE) (refer to Eq. (8)). In the case of classification, we have used binary cross entropy (refer to Eq. (9)).

$$E = \frac{1}{n} \sum_{k=1}^{np} (Y - V_k)^2 \tag{8}$$

$$E = \frac{1}{n}\sum_{k=1}^{np}(Y \cdot log(V_k) + (1-Y) \cdot log(1-V_k)) \quad (9)$$

After computing, the average weights were broadcasted to all the nodes. It is very important to note here that all of the partitions will receive identical parameters (i.e., *w, W, a,b*) in each iteration. The aggregation and then computing average is only done at a master node which demands the completion of the computing gradients at each data partition. In this way, the model is trained. Now, the same above process of computing gradients and then sending them off to master is continued for the user defined number of epochs, *N*. As mentioned earlier, the above training process is the same for both the regression and classification except for the error function.

After completing the above training process, the trained weights are utilized were used for the prediction on the test dataset and then results were reported. In the case of classification, sigmoid is used as discretizer to predict the class.

Algorithm 1: SPWNN algorithm for both classification and regression

**Input:** nin, nhn, non, w, W
**Output:** P : population evolved after mG, max generations

1: nin ← Input features
2: nhn ← Hidden nodes
3: non ← Output nodes
4: randomly initialize the required parameters w,W,a,b
5: bs ← batch size
6: **for all** machine {1...k} do in parallel
7:     for epoch 1 to N do
8:         w,W,a,b = mapper(batch) on each node (refer to Eq.(4)-(7))
            // weights are updated using SGD optimization
9:     **end for**
10:    Aggregate from all machines
11:    Compute average of w,W,a,b
12:    Compute Error
13: **end for**
14: Predict on Test data and report results

## 4.2 SPWNN for Streaming Data

In a streaming environment, the data is passed in as batches of streaming data which are obtained in a specified time interval. To handle such streaming data, SPWNN which is discussed in the previous section is not useful. Hence, we modified SPWNN to make it suitable for streaming data. Thus designed SPWNN can handle both the regression and classification analysis. The algorithm of SPWNN for streaming data is given in Algorithm 2.

Spark streaming context is initialized thereby injecting the streaming data. The entire data and partitioned into multiple batches and send as streams. Then by using Dstreams, an unbounded Spark data frame is created. In our approach, we have utilized the sliding window approach, where the data is collected in the form of batches. Each batch comprises a set of data points which are collected due to the course of the specified time interval. The number of batches to be collected is dependent on the sliding window length. These two parameters viz., sliding window length and time interval are the user defined parameters. Hence, these have to be initialized at the beginning. Suppose, the sliding window length is considered as *ws*, where the training is done on the first *ws*-1 batches and then testing is done on the latter batch of the sliding window. Since the length of the window is k, at a particular point in time, the window has atmost *ws* batches. If the window is not full (At least two batches are not streamed), then it will wait until it is full. On the other hand, if the window is full, then the k batches will be obtained from the window, training is done on the first *ws*-1 batches and test is performed on the last batch. After a certain time, or interval, the window slides by one batch. So, every time after sliding, prediction is done on the new data and training is done on the data that was predicted in the previous session. The training is performed on the first *ws*-1 batches of the window, where the weights are updated and the error is computed based on the underlying task i.e., in the case of classification (binary cross entropy is used), whereas in the case of regression (MSE is used). Using those weights the predictions were performed on the next batch which serves as the test batch.

Algorithm 2: SPWNN algorithm for streaming

```
1:  UnboundedTable ← Incoming Streams
2:  Initialize WindowSize (ws), Sliding Interval
3:  Start StreamingContext(Time)
4:  while IncomingStreams == True do
5:      UnboundedTable ← AppendIncomingDstreams
6:      If Window == Full then
7:          TrainData ← Window[:ws-1]
8:          TestData ← Window[ws]
9:      endif
10:     Slide window by one position across UnboundedTable
11: end while
12: Stop StreamingContext
```

## 5. Experiments & Results

### 5.1 Experimental Setup

All the experiments were performed in a Hadoop-Spark cluster comprising five nodes. All the Systems have the following configuration: Processor as Intel i5 8th generation, a RAM of 32 GB each with 4 physical and 8 logical cores and Ubuntu 18.04 Operating System. The versions of Hadoop is 2.7.3 and the Spark version is 2.2.0 [7] respectively.

### 5.2 Dataset Description

The datasets used in the current study are given in Table 1. Regression data sets are taken from UCI repository [29]. They contain time series obtained from 16 chemical sensors exposed to gas mixtures at different concentration levels. In particular, two gas mixtures are produced: Ethylene and Methane in the air and Ethylene and Carbon Monoxide in the air. For each measurement (each gas mixture), the signal is collected continuously for approximately 12 hours without interruption. In particular, the ethylene concentration varies between 0-20 ppm; CO is 0-600 ppm; and 0-300 ppm methane. the data is distributed in 19 columns. The first column represents the time (in seconds), the second column represents the methane (or CO) concentration set point (in ppm), the third column specifies the ethylene concentration set point (in ppm), then the next 16 columns show the records of the sensor array.

Classification datasets are taken from OpenML repository [34]. Such taken classification datasets are as follows: OVA Omentum and OVA Uterus are from GEMLeR, a collection of gene expression data sets, which can be used to compare gene expression-oriented machine learning algorithms. Every gene expression sample in these data sets comes from the large public repository expO (Expression Project for Oncology) of the International Genomics Consortium. Obtain tissue samples under standard conditions and perform gene expression analysis on a set of clinically annotated deidentified tumour samples. OVA Omentum and OVA Uterus have more than 10,000 features and Tissue is the Target variable for each dataset, which is a Categorical variable. The rest of the features were all numerical values only, where each column corresponds to values related to different factors affecting the Tissue.

Table 1: Details of the benchmark datasets

| Problem | Dataset | Instances | Features | Dependent Variables | Size |
|---|---|---|---|---|---|
| Regression | Ethylene Methane | 4178504 | 19 | Ethylene, Methane | 1.26 GB |
| Regression | Ethylene CO | 4208261 | 19 | Ethylene, CO | 1.31 GB |
| Classification | OVA Omentum | 1545 | 10936 | Tissue | 103 MB |
| Classification | OVA Uterus | 1545 | 10936 | Tissue | 106 MB |

**5.3 Data Pre-Processing**

For the Regression datasets, the time attribute from both files is dropped off. Thereafter, the data is normalized. The train-test split ratio is maintained as 80:20. Each regression dataset, contains two dependent variables. Currently, we are performing experiments on the multiple output single-output (MISO). Hence, each dataset is divided into two different datasets, wherein in a single case, one of the two dependent variables is used as a target while the other is used as a feature. Hence, Ethylene CO is named Ethylene CO – Ethylene where Ethylene is a dependent variable and Ethylene CO is named Etheylen CO – CO where the CO is a dependent variable. Similarly, Ethylene Methane is also named depending on the dependent variable used.

While, the Classification datasets, OVA Omentum and OVA Uterus are of high dimensional. Thereafter, the data is normalized. Same as previous, the train-test split ratio is 80:20. It's very important to note that, training with these many features didn't yield good results, so feature selection is performed and was needed. We have introduced the feature selection based on t-value statistics, where the features are sorted based on their t-value. The higher the t-value, the better the feature has to be selected. In our analysis, we have taken the top 100 features for the training classification model.

## 5.4 Classification Models

In this section, the results of the classification model in a static environment and on streaming data are discussed. The static results are presented in Table 3 and the streaming results are discussed in Table 4 to Table 7.

It turned out that overall, the SPWNN with Morlet as the activation performed well in the static environment. However, in streaming, SPWNN with Gaussian as the activation function outperformed the Morlet.

### 5.4.1 Classification results in Static Environment

Table 2: Best Hyperparameter combination for the classification dataset in static environment

| Hyperaparameter | Value |
|---|---|
| Number of hidden nodes (nhn) | 150 |
| Learning rate | 0.45 |
| Momentum | 0.999 |
| Batch size | 32 |
| Epochs | 100 |

After rigorous hyperparameter tuning the best performing combination is presented in Table 2. All the experiments were performed in the same cluster configuration as discussed in Section 5.2. It is observed that the training time consumed by SPWNN is almost similar irrespective of the activation function. The results after performing different activation functions are given in Table 3. The Area under Receiver Operator Characteristic Curve (AUC) measure is considered to choose the best model. Because, AUC is proven to be a robust measure when handling imbalanced datasets. SPWNN with Morlet as the activation function outperformed SPWNN with Gaussian in both datasets. The above analysis is observed since Morlet activation achieved a faster convergence than the Gaussian. Also, Morlet was able to learn the underlying hidden pattern better than the Gaussian which resulted in achieving better specificity. The same is observed in Fig A.1 and Fig A.2 where the graphs show how the SPWNN achieved sensitivity, specificity and AUC at various levels of epochs.

Table 3: Classification Results in Static Environment

| Dataset | A.F | Specificity | Sensitivity | AUC | Training Time (s) | Testing Time (s) |
|---|---|---|---|---|---|---|
| OVA Uterus | Morlet | 0.818792 | 0.986885 | **0.964186** | 2245.39 | 1.50 |
| OVA Uterus | Gaussian | 0.902685 | 0.908197 | 0.94199 | 2191.60 | 1.78 |
| OVA Omentum | Morlet | 0.94586 | 0.874194 | **0.957791** | 2215.45 | 2.99 |
| OVA Omentum | Gaussian | 0.812081 | 0.990164 | 0.955382 | 2181.52 | 2.72 |

* where A.F is activation function

### 5.4.2 Classification results on Streaming Environment

The best performing hyperparameter combination for SPWNN for streaming is presented in Table 4. The window size is chosen as 2. Hence, there are two different stream batches in each window. Here, the training is performed on the former stream batch and the testing is performed on the latter stream batch respectively. The total number of stream batches was fixed to be 10 in our experiment analysis. The results of OVA Omentum where the activation function (AF) is Morlet are discussed in Table 5 and results with respect to Gaussian AF are discussed in Table 6. The results were given for every batch to show the behaviour of the proposed streaming model.

Table 4: Best Hyperparameter combination for the classification dataset in streaming environment

| Hyperaparameter | Value |
|---|---|
| Number of hidden nodes (nhn) | 150 |
| Learning rate | 0.2 |
| Momentum | 0.999 |
| Batch size | 16 |
| Epochs | 100 |

Here also, the training time is almost similar for both the AFs. The results after performing with different activation functions are given in Table 5 and Table 6 for the OVA Uterus dataset and the results corresponding to OVA Omentum with Morlet are given in Table 7 and Gaussian in Table 8. As discussed earlier, there are 9 sliding windows, and the average AUC obtained over them is considered for evaluating the best model. In the streaming environment, unlike in a static environment, SPWNN with Gaussian outperformed SPWNN with Morlet. This owes to the fact that Gaussian is able to learn the underlying hidden pattern better than the Morlet. Even while comparing window wise, Gaussian outperformed Morlet in terms of AUC in majority of the windows in both the datasets. In the case of OVA Uterus dataset, SPWNN with Gaussian achieved almost similar AUC when compared to SPWNN with Morlet. However, it achieved significantly better in the case of the OVA Omentum dataset.

Table 5: Results obtained by SPWNN on OVA Uterus with A.F: Morlet in streaming environment

| Streaming Batches | Specificity | Sensitivity | AUC |
|---|---|---|---|
| Batch 1 | 0.96692125 | 0.73943662 | 0.97183099 |
| Batch 2 | 0.95858957 | 0.92253521 | 0.81690141 |
| Batch 3 | 0.96245785 | 0.94366197 | 0.87323944 |
| Batch 4 | 0.98031145 | 0.78169014 | 0.99295775 |
| Batch 5 | 0.94544733 | 0.84507042 | 0.93661972 |
| Batch 6 | 0.97361635 | 0.8943662 | 0.95774648 |
| Batch 7 | 0.95129935 | 0.77464789 | 0.96478873 |
| Batch 8 | 0.95065463 | 0.92957747 | 0.80985916 |
| Batch 9 | 0.9360246 | 0.8028169 | 0.97183099 |
| **Average** | **0.958369150** | **0.848200312** | **0.921752741** |

Table 6: Results obtained by SPWNN on OVA Uterus with A.F: Gaussian in streaming environment

| Streaming Batches | Specificity | Sensitivity | AUC |
|---|---|---|---|
| Batch 1 | 0.95427494 | 0.61267606 | 0.99145675 |
| Batch 2 | 0.93746281 | 0.66901409 | 0.97183099 |
| Batch 3 | 0.9188653 | 0.78169014 | 0.95774648 |
| Batch 4 | 0.93984329 | 0.78169014 | 0.96478873 |
| Batch 5 | 0.85553462 | 0.78873239 | 0.88732394 |
| Batch 6 | 0.90909542 | 0.87323944 | 0.91549296 |
| Batch 7 | 0.89565562 | 0.85915493 | 0.87323944 |
| Batch 8 | 0.92030351 | 0.85915493 | 0.85211268 |
| Batch 9 | 0.92615552 | 0.8028169 | 0.87323944 |
| **Average** | **0.91746567** | **0.780907668** | **0.92080349** |

Table 7: Results obtained by SPWNN on OVA Omentum with A.F: Morlet in streaming environment

| Streaming Batches | Specificity | Sensitivity | AUC |
|---|---|---|---|
| Batch 1 | 0.90258688 | 0.91156463 | 0.79591837 |
| Batch 2 | 0.73316674 | 0.59863946 | 0.78911565 |
| Batch 3 | 0.93641538 | 0.95238095 | 0.68707483 |
| Batch 4 | 0.95839696 | 0.93877551 | 0.86394558 |
| Batch 5 | 0.94506918 | 0.88435374 | 0.86394558 |
| Batch 6 | 0.9080013 | 0.93877551 | 0.79591837 |
| Batch 7 | 0.9494655 | 0.94557823 | 0.82312925 |
| Batch 8 | 0.92179583 | 0.62328767 | 0.93150685 |
| Batch 9 | 0.94234378 | 0.84931507 | 0.89726027 |
| **Average** | **0.910804618** | **0.849185641** | **0.827534971** |

Table 8: Results obtained by SPWNN on OVA Omentum with AF: Gaussian in streaming environment

| Streaming Batches | Specificity | Sensitivity | AUC |
|---|---|---|---|
| Batch 1 | 0.91026887 | 0.91156463 | 0.82312925 |
| Batch 2 | 0.8741728 | 0.80952381 | 0.89795918 |
| Batch 3 | 0.94145958 | 0.93877551 | 0.83673469 |
| Batch 4 | 0.95390809 | 0.86394558 | 0.95238095 |
| Batch 5 | 0.96001666 | 0.8707483 | 0.91836735 |
| Batch 6 | 0.91286038 | 0.95918367 | 0.82993197 |
| Batch 7 | 0.96233051 | 0.91156463 | 0.89115646 |
| Batch 8 | 0.9392006 | 0.89041096 | 0.86986301 |
| Batch 9 | 0.96631638 | 0.85616438 | 0.93150685 |
| **Average** | **0.935614874** | **0.890209051** | **0.8834477474** |

### 5.5 Regression Models

In this section, the results of the regression model in a static environment and on streaming data are discussed. The static results are presented in Table 10 and the streaming results are discussed in Table 11 to Table 14.

It turned out that overall, the SPWNN with Gaussian performed better than the SPWNN with Morlet in most of the cases in the static environment. However, in streaming, SPWNN with Gaussian as the activation function outperformed the Morlet.

### 5.5.1 Regression results on Static Environment

Table 9: Best Hyperparameter combination for the regression dataset in static environment

| Hyperaparameter | Value |
|---|---|
| Number of hidden nodes (nhn) | 10 |
| Learning rate | 0.45 |
| Momentum | 0.999 |
| Batch size | 2048 |
| Epochs | 1000 |

After rigorous hyperparameter tuning the best performing combination is presented in Table 9. All the experiments were performed in the same cluster configuration as discussed in Section 5.2. It is observed that the training time consumed by SPWNN is almost similar irrespective of the activation function. The results after performing with different activation functions are given in Table 10.

      In the case of regression, mean squared error (MSE) is considered the metric to decide the best model. The results indicate that except in the case of Ethylene CO, where Ethylene is considered as the prediction variable, SPWNN with Gaussian as the activation function outperformed SPWNN with Morlet. The above analysis is observed owing to the fact that Gaussian activation achieved a faster convergence than the Morlet. Also, Gaussian can learn the underlying hidden pattern better than the Morlet which resulted in achieving lower MSE. The same is observed in Fig A.1 and Fig A.2 where the graphs show how the SPWNN achieved MSE at various levels of epochs.

Table 10: Regression results obtained by SPWNN in static environment

| Dataset | Prediction variable | A F | MSE on Test | Training Time (s) | Testing Time (s) |
|---|---|---|---|---|---|
| Ethylene CO | CO | Morlet | 0.055885709 | 11458.69 | 23.69 |
| Ethylene CO | CO | Gaussian | **0.045988779** | 11669.34 | 39.45 |
| Ethylene CO | Ethylene | Morlet | **0.044376956** | 11772.71 | 16.79 |
| Ethylene CO | Ethylene | Gaussian | 0.062724334 | 11521.48 | 46.93 |
| Ethylene Methane | Methane | Morlet | 0.037892088 | 11656.79 | 26.64 |
| Ethylene Methane | Methane | Gaussian | **0.028004120** | 11156.69 | 33.46 |
| Ethylene Methane | Ethylene | Morlet | 0.046449774 | 11181.62 | 43.39 |
| Ethylene Methane | Ethylene | Gaussian | **0.039013186** | 11564.88 | 44.93 |

### 5.5.2 Regression results on Streaming Environment

Table 11: Best Hyperparameter combination for the regression dataset in streaming environment

| Hyperaparameter | Value |
|---|---|
| Number of hidden nodes (nhn) | 10 |
| Learning rate | 0.20 |
| Momentum | 0.999 |
| Batch size | 512 |
| Epochs | 100 |

The best performing hyperparameter combination for SPWNN for streaming is presented in Table 11. The window size is chosen as 2. Hence, there are two different stream batches in each window. Here, the training is performed on the former stream batch and the testing is performed on the latter stream batch respectively. The total number of stream batches was fixed to be 20 in our experiment analysis. The results of Ethylene Methane- Ethylene, where the activation function (AF) is Morlet are discussed in Table 12 and results with respect to Gaussian AF, are discussed in Table 13. Similarly, results corresponding to Ethylene Methane – Methane were discussed in Table 14 (Morlet as AF) and Table 15 (Gaussian as AF) respectively. Also, Ethylene CO – Etheylene were discussed in Table 16 and Table 17 where Morlet and Gaussian as AFs. Similarly, Ethylene CO – CO were discussed in Table 18 (Morlet as AF) and Table 19 (Gaussian as AF) respectively. The results were given for every batch to show the behaviour of the proposed streaming model.

Here also, the training time is almost similar for both the AFs. As discussed earlier, there are 19 sliding windows, and the average MSE obtained over them is considered for evaluating the best model. In the streaming environment also, SPWNN with Gaussian outperformed SPWNN with Morlet in the majority of the cases except in Ethylene CO - CO. This owes to the fact that Gaussian can learn the underlying hidden pattern better than Morlet. While comparing window wise also, Gaussian outperformed Morlet in terms of MSE in the majority of the datasets in the majority of the windows. The above analysis is observed owing to the fact that Gaussian activation achieved a faster convergence than the Morlet. Also, Gaussian was able to learn the underlying hidden pattern better than Morlet which resulted in achieving lower MSE.

| Table 12: Results obtained by SPWNN on Ethylene Methane - Ethylene, AF: Morlet in streaming environment | |
|---|---|
| **Streaming Batches** | **Test MSE** |
| Batch 1 | 0.12393604 |
| Batch 2 | 0.07378749 |
| Batch 3 | 0.07004135 |
| Batch 4 | 0.19466037 |
| Batch 5 | 0.09869748 |
| Batch 6 | 0.09127234 |
| Batch 7 | 0.05604061 |
| Batch 8 | 0.12436425 |
| Batch 9 | 0.12374341 |
| Batch 10 | 0.12379226 |
| Batch 11 | 0.12349314 |
| Batch 12 | 0.124275 |
| Batch 13 | 0.12370406 |
| Batch 14 | 0.12352001 |
| Batch 15 | 0.12432016 |
| Batch 16 | 0.12404439 |
| Batch 17 | 0.12378456 |
| Batch 18 | 0.12389296 |
| Batch 19 | 0.12351377 |
| **Average** | **0.1155201928** |

| Table 13: Results obtained by SPWNN on Ethylene Methane - Ethylene, AF: Gaussian in streaming environment | |
|---|---|
| **Streaming Batches** | **Test MSE** |
| Batch 1 | 0.04954237 |
| Batch 2 | 0.0692704 |
| Batch 3 | 0.04018015 |
| Batch 4 | 0.06255346 |
| Batch 5 | 0.12202037 |
| Batch 6 | 0.12435347 |
| Batch 7 | 0.12410548 |
| Batch 8 | 0.12427145 |
| Batch 9 | 0.05127125 |
| Batch 10 | 0.04764669 |
| Batch 11 | 0.09311369 |
| Batch 12 | 0.05625233 |
| Batch 13 | 0.07235238 |
| Batch 14 | 0.12344497 |
| Batch 15 | 0.12431177 |
| Batch 16 | 0.05355372 |
| Batch 17 | 0.05513086 |
| Batch 18 | 0.05183397 |
| Batch 19 | 0.068624 |
| **Average** | **0.079675408021** |

Table 14: Results obtained by SPWNN on Ethylene Methane - Methane, AF: Morlet in streaming environment

| Streaming Batches | Test MSE |
|---|---|
| Batch 1 | 0.11095944 |
| Batch 2 | 0.11099098 |
| Batch 3 | 0.10724215 |
| Batch 4 | 0.12713889 |
| Batch 5 | 0.11061057 |
| Batch 6 | 0.10502671 |
| Batch 7 | 0.10616078 |
| Batch 8 | 0.10557956 |
| Batch 9 | 0.10466019 |
| Batch 10 | 0.10560901 |
| Batch 11 | 0.10550903 |
| Batch 12 | 0.10487179 |
| Batch 13 | 0.10503021 |
| Batch 14 | 0.1069103 |
| Batch 15 | 0.09554048 |
| Batch 16 | 0.0587573 |
| Batch 17 | 0.05695155 |
| Batch 18 | 0.10409125 |
| Batch 19 | 0.1044101 |
| **Average** | **0.10189738314** |

Table 15: Results obtained by SPWNN on Ethylene Methane - Methane, AF: Gaussian in streaming environment

| Streaming Batches | Test MSE |
|---|---|
| Batch 1 | 0.2036292 |
| Batch 2 | 0.16782641 |
| Batch 3 | 0.09504617 |
| Batch 4 | 0.17521183 |
| Batch 5 | 0.20498493 |
| Batch 6 | 0.0845818 |
| Batch 7 | 0.05578543 |
| Batch 8 | 0.11388623 |
| Batch 9 | 0.12824928 |
| Batch 10 | 0.07030886 |
| Batch 11 | 0.11057301 |
| Batch 12 | 0.09306753 |
| Batch 13 | 0.09478232 |
| Batch 14 | 0.02507607 |
| Batch 15 | 0.05470719 |
| Batch 16 | 0.03640039 |
| Batch 17 | 0.01953907 |
| Batch 18 | 0.04297256 |
| Batch 19 | 0.0633842 |
| **Average** | **0.0814470302975433** |

Table 16: Results obtained by SPWNN on Ethylene CO - Ethylene, AF: Morlet in streaming environment

| Streaming Batches | Test MSE |
|---|---|
| Batch 1 | 0.1521719 |
| Batch 2 | 0.17107978 |
| Batch 3 | 0.15176993 |
| Batch 4 | 0.15144301 |
| Batch 5 | 0.14428568 |
| Batch 6 | 0.12974817 |
| Batch 7 | 0.09812364 |
| Batch 8 | 0.08547947 |
| Batch 9 | 0.06285561 |
| Batch 10 | 0.09247555 |
| Batch 11 | 0.06300222 |
| Batch 12 | 0.07052327 |
| Batch 13 | 0.15408002 |
| Batch 14 | 0.06293924 |
| Batch 15 | 0.06318491 |
| Batch 16 | 0.08479652 |
| Batch 17 | 0.16855028 |
| Batch 18 | 0.11286866 |
| Batch 19 | 0.05984846 |
| **Average** | **0.1094329642** |

Table 17: Results obtained by SPWNN on Ethylene CO - Ethylene, AF: Gaussian in streaming environment

| Streaming Batches | Test MSE |
|---|---|
| Batch 1 | 0.15039672 |
| Batch 2 | 0.15245786 |
| Batch 3 | 0.14977086 |
| Batch 4 | 0.14955621 |
| Batch 5 | 0.14797773 |
| Batch 6 | 0.14970488 |
| Batch 7 | 0.14931455 |
| Batch 8 | 0.05991966 |
| Batch 9 | 0.06808535 |
| Batch 10 | 0.0878329 |
| Batch 11 | 0.04406603 |
| Batch 12 | 0.06351105 |
| Batch 13 | 0.06830484 |
| Batch 14 | 0.05413749 |
| Batch 15 | 0.05023188 |
| Batch 16 | 0.08657154 |
| Batch 17 | 0.05751673 |
| Batch 18 | 0.07321593 |
| Batch 19 | 0.04820837 |
| **Average** | **0.0953042413** |

Table 18: Results obtained by SPWNN on Ethylene Methane - CO, AF: Morlet in streaming environment

| Streaming Batches | Test MSE |
|---|---|
| Batch 1 | 0.15200634 |
| Batch 2 | 0.15267159 |
| Batch 3 | 0.14880514 |
| Batch 4 | 0.14860076 |
| Batch 5 | 0.08102283 |
| Batch 6 | 0.24311809 |
| Batch 7 | 0.11596852 |
| Batch 8 | 0.11308452 |
| Batch 9 | 0.10755804 |
| Batch 10 | 0.11518632 |
| Batch 11 | 0.08722476 |
| Batch 12 | 0.10341598 |
| Batch 13 | 0.07294112 |
| Batch 14 | 0.06055401 |
| Batch 15 | 0.14785037 |
| Batch 16 | 0.14714199 |
| Batch 17 | 0.14714337 |
| Batch 18 | 0.14809651 |
| Batch 19 | 0.14717899 |
| **Average** | **0.1283983551** |

Table 19: Results obtained by SPWNN on Ethylene Methane - CO, AF: Gaussian in streaming environment

| Streaming Batches | Test MSE |
|---|---|
| Batch 1 | 0.14918636 |
| Batch 2 | 0.14883504 |
| Batch 3 | 0.14758017 |
| Batch 4 | 0.14721954 |
| Batch 5 | 0.10613069 |
| Batch 6 | 0.30087736 |
| Batch 7 | 0.0927142 |
| Batch 8 | 0.11539932 |
| Batch 9 | 0.10297482 |
| Batch 10 | 0.08172066 |
| Batch 11 | 0.09331961 |
| Batch 12 | 0.09650104 |
| Batch 13 | 0.07961465 |
| Batch 14 | 0.10225569 |
| Batch 15 | 0.14756432 |
| Batch 16 | 0.14788124 |
| Batch 17 | 0.14783506 |
| Batch 18 | 0.14871924 |
| Batch 19 | 0.14799978 |
| **Average** | **0.13180677866983** |

## 5.7 Speedup analysis

Scalability is used as a measure which reflects the ability of the system to effectively use resources. Speedup is an indicator to verify our parallel architecture's performance, it is defined as the relationship between the serial execution time of the best sequential algorithm to solve the problem and the time it takes for the parallel algorithm to solve the same problem in p processors. The mathematical formulation of Speed up is given in Eq. (10).

$$Speed\ Up = \frac{T_S}{T_P} \qquad (10)$$

Speedup achieved by the proposed model is given in Table 20. The time taken by the SPWNN in both the datasets is almost similar. Further, the effect of change in activation function is also observed to be nothing. Hence, the speed up analysis is conducted on only one dataset in each machine learning task i.e., in classification and regression. It is observed that proposed SPWNN achieved speedup of 1.32-1.40.

Table 20: Speed up Results

| Model | Sequential version execution time | SPWNN execution time | Speed Up |
|---|---|---|---|
| Regression Model | 15673.45 | 11156.69 | 1.40 |
| Classification Model | 1512.67 | 1142.21 | 1.32 |

## 6. Conclusions

In summary, we developed a Scalable and Parallelized Wavelet Neural Network (SPWNN), under the Apache Spark environment to solve both regression and classification problems. To achieve horizontal parallelization, we employed parallel SGD algorithm to train the SPWNN. The same data was used in the streaming environment with the help of Queue Streams in Spark Streaming to handle big data effectively. As for classification, Morlet worked better for OVA Omentum and OVA Uterus datasets. However, while streaming, the Average MSE error obtained by SPWNN with Gaussian activation function turned out to be less than that of Morlet activation function in the majority of the cases.

The proposed SPWNN can further be extended for the multi-layer SPWNN to enable to learn more complex patterns thereby making it Scalable and Parallelized Deep Wavelet Neural Network (SPDWNN). In future, we would like to explore various other wavelet transforms which could be used as activation functions in different hidden layers to achieve better results with a small number of hidden nodes. Streaming SPWNN model can further be extended to solve nowcasting the GDP and other such financial time series etc.

# Appendix

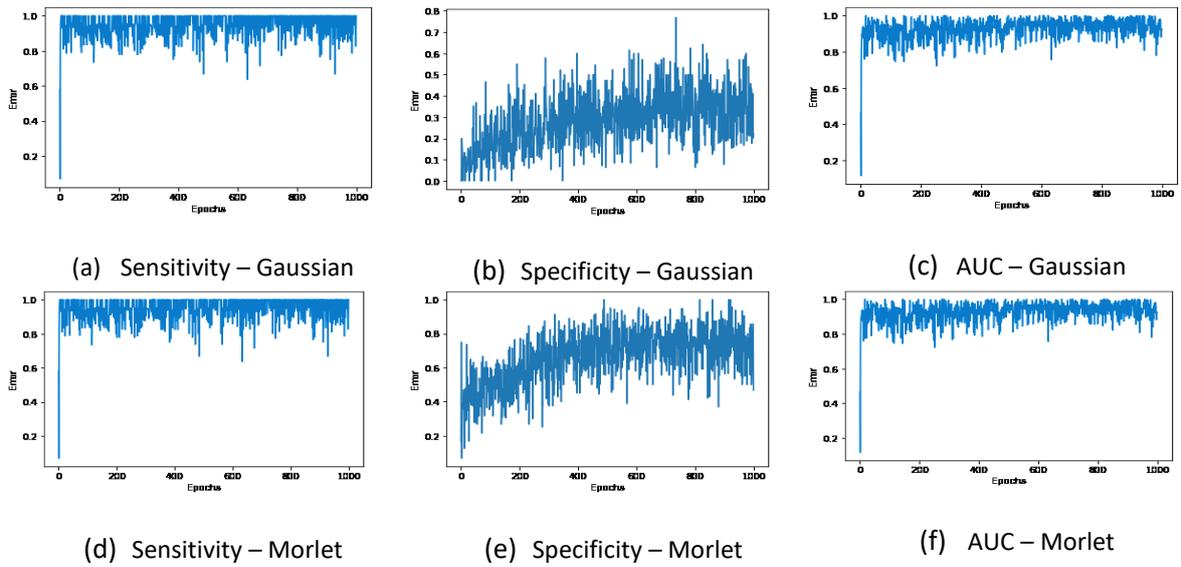

Fig. A.1. Classification Results on OVA Omentum Dataset

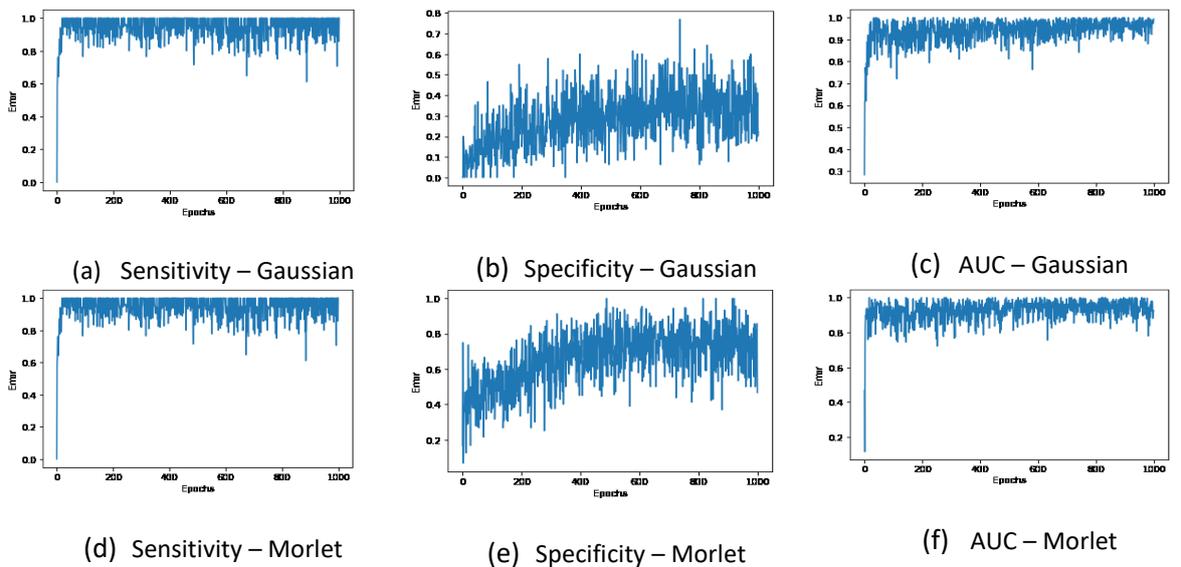

Fig. A.2. Classification Results on OVA Uterus Dataset

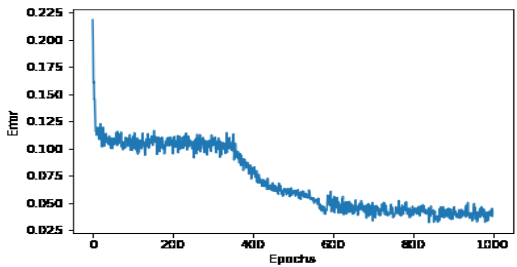
(a) Methane Concentration - Morlet

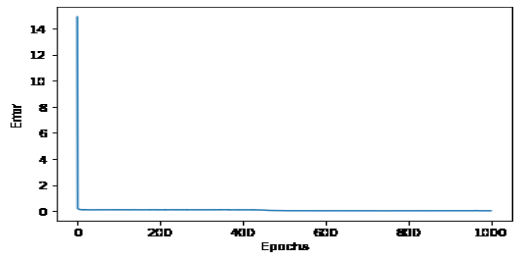
(b) Methane Concentration - Gaussian

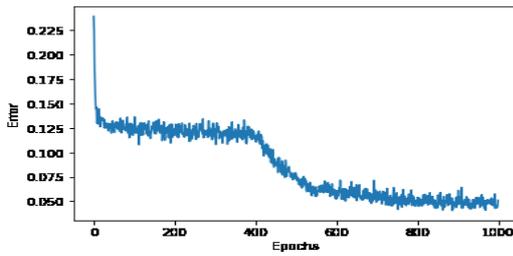
(c) Ethylene Concentration - Morlet

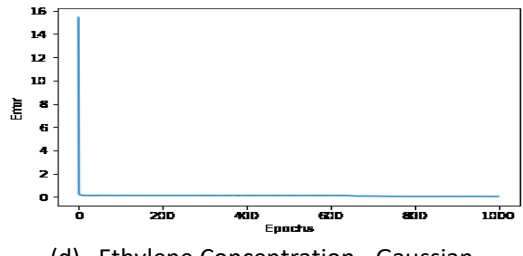
(d) Ethylene Concentration - Gaussian

Fig. A.3. Regression Results on Ethylene Methane Dataset

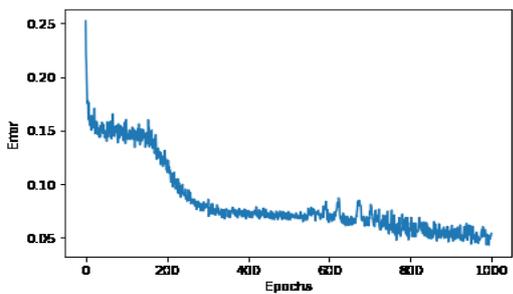
(a) CO Concentration - Morlet

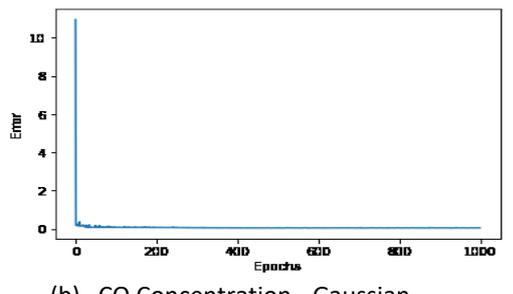
(b) CO Concentration - Gaussian

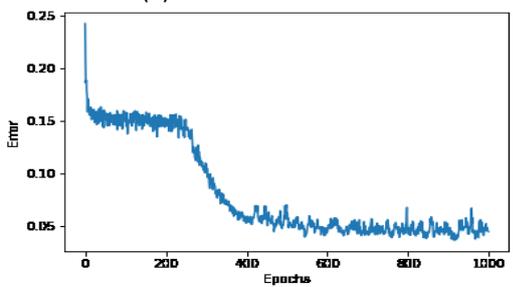
(c) Ethylene Concentration - Morlet

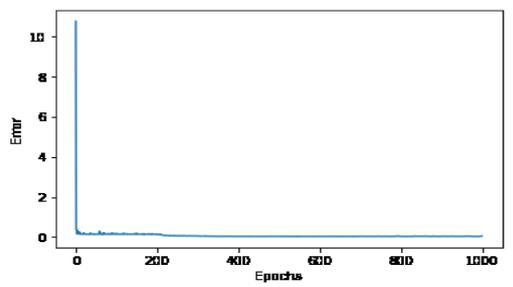
(d) Ethylene Concentration - Gaussian

Fig. A.4. Regression Results on Ethylene CO Dataset